\documentclass[11pt]{article}

\usepackage{blindtext}
\usepackage[utf8]{inputenc}
\usepackage{natbib}
\usepackage{graphicx}
\usepackage{rotating}
\usepackage{pdflscape}
\usepackage{url}
\usepackage{amsmath}
\usepackage{multirow}
\usepackage{lipsum}       
\usepackage{changepage}
\usepackage{enumitem}
\usepackage{color}
\usepackage{inputenc}
\usepackage{pgfplots}

\renewcommand{\theenumii}{\theenumi.\arabic{enumii}.}
\begin{document}

\title{PQuAD: A \textbf{P}ersian  \textbf{Qu}estion  \textbf{A}nswering  \textbf{D}ataset}


\author{Kasra Darvishi, Newsha Shahbodagh, Zahra Abbasiantaeb, Saeedeh Momtazi*	\\
Department of Computer Engineering \\
Amirkabir University of Technology (Tehran Polytechnic) \\
}



\date{}

\maketitle

\begin{abstract}
We present Persian Question Answering Dataset (PQuAD), a crowdsourced reading comprehension dataset on Persian Wikipedia articles. It includes 80,000 questions along with their answers, with 25\% of the questions being adversarially unanswerable. We examine various properties of the dataset to show the diversity and the level of its difficulty as a MRC benchmark. By releasing this dataset, we aim to ease research on Persian reading comprehension and development of persian question answering systems. Our experiments on different state-of-the-art pre-trained contextualized language models shows 74.8\% Exact Match (EM) and 87.6\% F1-score that can be used as the baseline results for further research on Persian QA.


\end{abstract}

\section{Introduction}

Machine Reading Comprehension (MRC) is one of the central tasks in natural language understanding which requires a system to read a passage and then answer the given questions from the passage. Developing a machine that has the comprehension ability helps Artificial Intelligence (AI) to reach the goal of competing with human. Developing such a machine that can comprehend the available data in the text format would be highly beneficial, because in many jobs human workers are required to get information from text and having a high performance MRC model will help to reduce manual tasks significantly.

MRC has undergone significant advancements in recent years due to the availability of many large-scale annotated datasets, including MCTest \citep{richardson:2013:mctest}, BookTest \citep{bajgar:2016:booktest}, SQuAD \citep{rajpurkar:2016:squad}, SearchQA \citep{dunn:2017:searchqa}, NewsQA \citep{Trischler:2017:newsqa}, ReCoRD \citep{zhang:2018:record} and ReCO \citep{wang:2020:reco}. These datasets provide the opportunity to train data-intensive deep learning models and achieve performance comparable to humans. 

The main MRC datasets, however, are particularly in English, while low-resource languages such as Persian require further efforts in this regard. Recently, some progress has been made towards this goal, namely PersianQA \citep{PersianQA} and ParSQuAD \citep{abadani:2021}, which are gathered automatically by different methodologies. However, they either lack the sufficient number of examples to be used in a supervised learning framework for today’s large models or are not as high quality as manual ones.
ParSQuAD \citep{abadani:2021} is collected by translating the SQuAD \citep{rajpurkar:2016:squad} dataset into Persian by google-translate and PersianQA \citep{PersianQA} is a small size MRC dataset consisting of 10,000 questions, which is not suitable for training the large models. 

In this paper, we introduce PQuAD, a large-scale and high quality Persian span-extraction MRC dataset consisting of 80,000 human-annotated questions. PQuAD’s questions are based on Persian Wikipedia articles and cover a wide variety of subjects. The dataset has been used for training the state-of-the-art models in the field.

The rest of the paper is organized as follows: in Section 2, we briefly overview related MRC datasets. Section 3 describes the data collection process of PQuAD and methodologies used to ensure a high quality dataset. In Section 4 we analyze various properties of the dataset to show the level of its difficulty as a MRC benchmark. Section 5 is devoted to introduction of baseline models and their results on PQuAD. Finally, Section 6 concludes the paper.

\section{Related Works}
\label{sec:relatedWork}
The increasing attention of the researchers toward MRC has enhanced the importance of providing the MRC datasets. Although a large number of MRC datasets are available \citep{richardson:2013:mctest,bajgar:2016:booktest,rajpurkar:2016:squad,dunn:2017:searchqa,Trischler:2017:newsqa,zhang:2018:record,wang:2020:reco}, it is still important to collect a MRC with new features. 

Recently, there has been a great motivation toward collecting the MRC datasets for low-resource languages  \citep{mozannar:2019:arabic,carrino:2020:spanish,dhoffschmidt:2020:fquad,lim:2019:korquad1}. The availability of the multi-lingual pre-trained language models, such as Multilingual BERT \citep{devlin:2019:bert}, XLM-RoBERTa \citep{conneau:2020:xlmroberta}, can be considered as one of the motivations behind this motivation. 
Most of these datasets, are collected from wikipedia pages similar to SQuAD or are curated by translating the SQuAD dataset.

MCTest \citep{richardson:2013:mctest} is one of the earliest MRC datasets which includes fictional stories and multiple choice questions collected for each story. This dataset is very small and it is not sufficient for training large models.

SQuAD 1.1 \citep{rajpurkar:2016:squad} is a large-scale MRC dataset which is collected from wikipedia pages. Answer of each question is represented as a span of text within the corresponding passage. In SQuAD 2.0, more than 50,000 unanswerable questions are added. The unanswerable questions are very similar to the answerable questions and this feature makes the SQuAD 2.0 dataset a more challenging dataset. Nevertheless, the available studies on MRC using SQuAD 2.0 \citep{yamada:2020,yang:2019,joshi:2020:spanbert} have achieved a great performance over human evaluation. 

NewsQA \citep{Trischler:2017:newsqa} is a large-scale dataset which includes 119,633 questions gathered by human workers from 12,744 CNN's news articles. Among the available studies on NewsQA dataset \citep{joshi:2020:spanbert,tay:2018,kundu:2018}, the SpanBERT model \citep{joshi:2020:spanbert} has achieved a higher performance compared to the human judgment. 

SearchQA \citep{dunn:2017:searchqa} is collected according to the procedure of retrieving the correct answer in a real question answering system. Different from other datasets which a question is posed given the passage, the question is posed and then the relevant passages are retrieved for the given question. 
SearchQA includes more than 140000 question-answer pairs and each question-answer pair is linked to about 50 text snippets.

ReCoRD \citep{zhang:2018:record} is collected automatically from news articles. The distinctive feature of this dataset is that despite the other MRC datasets, such as SQuAD and NewsQA, it needs commonsense reasoning over several sentences for comprehending the answer. A small portion of the questions of this dataset can be answered by paraphrasing. 

There are several MRC datasets for other languages than English, including ReCo \citep{wang:2020:reco} in Japanese, ParSQuAD \citep{abadani:2021} in Persian, FQuAD \citep{d:2020:fquad} in French, ARCD \citep{mozannar:2019} in Arabic, KorQuAD1.0 \citep{lim:2019:korquad1} in Korea, and Spanish translation of SQuAD \citep{carrino:2020}.

ARCD \citep{mozannar:2019} consists of Arabic translation of SQuAD and questions posed by crowd workers. ParSQuAD \citep{abadani:2021} in Persian is also gathered by translating the SQuAD into Persian. For Spanish SQuAD \citep{carrino:2020}, a trained neural machine translation and a trained unsupervised word alignment model are developed for automatically translating the SQuAD dataset to Spanish. FQuAD \citep{d:2020:fquad} is collected from French Wikipedia pages. ReCO \citep{wang:2020:reco} is a very large opinion-based MRC dataset including 300,000 Japanese questions which includes both factoid and non-factoid questions.

Another type of the datasets is cloze datasets. In the cloze datasets, a word is omitted from the text and goal is to detect the omitted word from the given text. Children’s Book test (CBT) \citep{hill:2015} dataset is a cloze dataset. Each sample in this dataset is 21 consecutive sentences and one word in the last sentence is omitted. The first 20 sentences are given as context and the missing word in the next sentence must be predicted.
BookTest \citep{bajgar:2016:booktest} dataset is very similar to the CBT dataset but it is designed for training large models due to its 60 times larger size than CBT.

\section{Dataset Collection}
\label{sec:data-collection}

The data collection process of PQuAD consists of three stages: passage curation, question-answer pair annotation, and additional answer collection.

\subsection{Passage curation}
The most important articles in the Persian Wikipedia are selected based on two criteria: (1) having an info-box, and (2) being among the top pages in the Persian Wikipedia based on the PageRank algorithm. 

We believe that important Wikipedia pages, that have enough content for question collection, usually contain an info-box and we use it as the primary condition to retrieve 600,000 articles. 

In the next step, we build a graph for these articles based on their links and rank the collected articles using the NetworkX library’s\footnote{https://networkx.org/} PageRank method with damping parameter set to 0.4. Having the obtained set of ordered articles, for each article, we choose the introduction section and the first 20 paragraphs, whose length is 500-1100 characters. The total number of paragraphs is 11,000, which belong to 1,125 articles spanning a wide range of topics.

\subsection{Question-answer annotation}
Crowdworkers were instructed to spend 15 minutes on each paragraph and collect about five to ten questions as well as their answers, if they are answerable using the paragraph's content. At the same time, they discarded the paragraphs that were hard to understand or contained too many non-Persian words. To further regulate the process, we developed a toolkit for the annotation process. In this toolkit, crowdworkers typed each question in a text box and then either specified an answer by marking a span or labeled the question as unanswerable. Workers were also encouraged to use paraphrased sentences in their questions and avoid choosing the answers comprising non-Persian words. 
Unanswerable questions roughly form 25\% of all questions and were collected in such a way that are in the same content of the paragraph, but they could not be answered just by reading the paragraph.

\subsection{Additional answer collection}

To have a better evaluation of models, we further check the question-answer pairs in the test and validation sets while additional answers are being collected. A new group of crowdworkers answer the questions in these two sets. The crowdworkers were encouraged to include the other correct answer spans with minor differences to the original answers. Moreover, if they find that the original answer is incorrect, they resolved the incorrect answer either by choosing the more appropriate answer or removing the corresponding question from the passage.  The distribution of the number of answers per question is represented in Figure \ref{fig:ques-dist-1}.

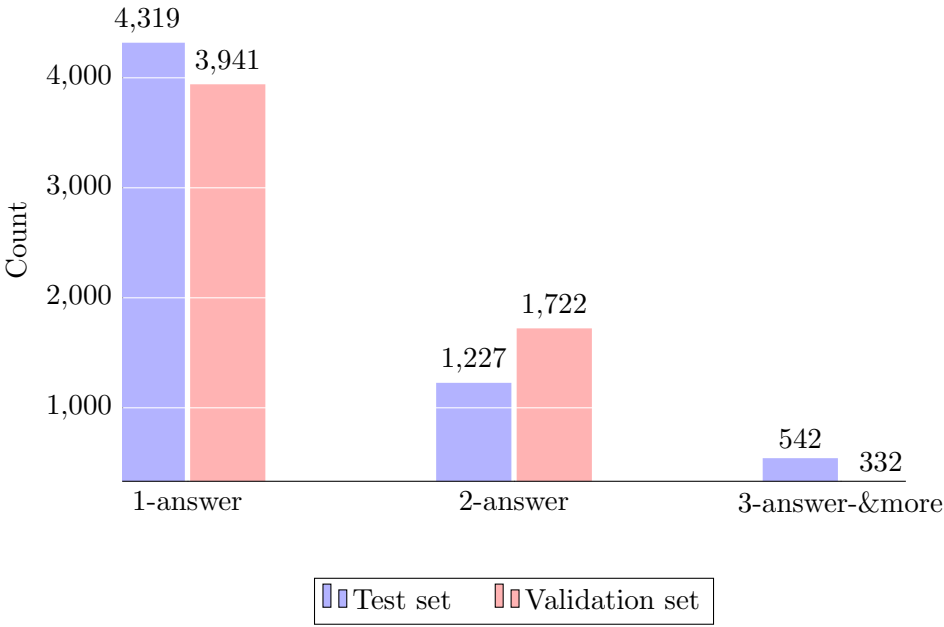
\begin{figure*}
	\caption{Distribution of number of answers available for each answerable question in test and validation sets.}
	\label{fig:ques-dist-1}
	\pgfplotsset{width=8cm,compat=1.8}
	\renewcommand*{\familydefault}{\sfdefault}
	\begin{tikzpicture}
		\label{fig:ques-dist-1}
		\centering
		\begin{axis}[
			ybar, axis on top,
			height=8cm, width=12cm,
			bar width=1cm,
			ymajorgrids, tick align=inside,
			major grid style={draw=white},
			enlarge y limits={value=.1,upper},
			axis x line*=bottom,
			axis y line*=left,
			y axis line style={opacity=0},
			tickwidth=0pt,
			enlarge x limits=true,
			legend style={
				at={(0.5,-0.2)},
				anchor=north,
				legend columns=-1,
				/tikz/every even column/.append style={column sep=0.5cm}
			},
			ylabel={Count},
			symbolic x coords={
				1-answer,2-answer,3-answer-\&more},
			xtick=data,
			nodes near coords={
				\pgfmathprintnumber[precision=0]{\pgfplotspointmeta}
			}
			]
			\addplot [draw=none, fill=blue!30] coordinates {
				(1-answer,4319)
				(2-answer, 1227) 
				(3-answer-\&more,542) };
			
			\addplot [draw=none,fill=red!30] coordinates {
				(1-answer,3941)
				(2-answer, 1722) 
				(3-answer-\&more,332)};
			
			\legend{Test set, Validation set}
		\end{axis}
	\end{tikzpicture}
\end{figure*}


\section{Dataset Analysis}
\label{sec:ds-analysis}
We investigate PQuAD in order to demonstrate the level of its challenge and its various properties. Types of answers and some statistics about the different domains available in the dataset, as well as general information, are included in this analysis.

\subsection{Dataset Structure}
\label{sec:ds-structure}
\begin{figure*}
	\includegraphics[width=1\textwidth]{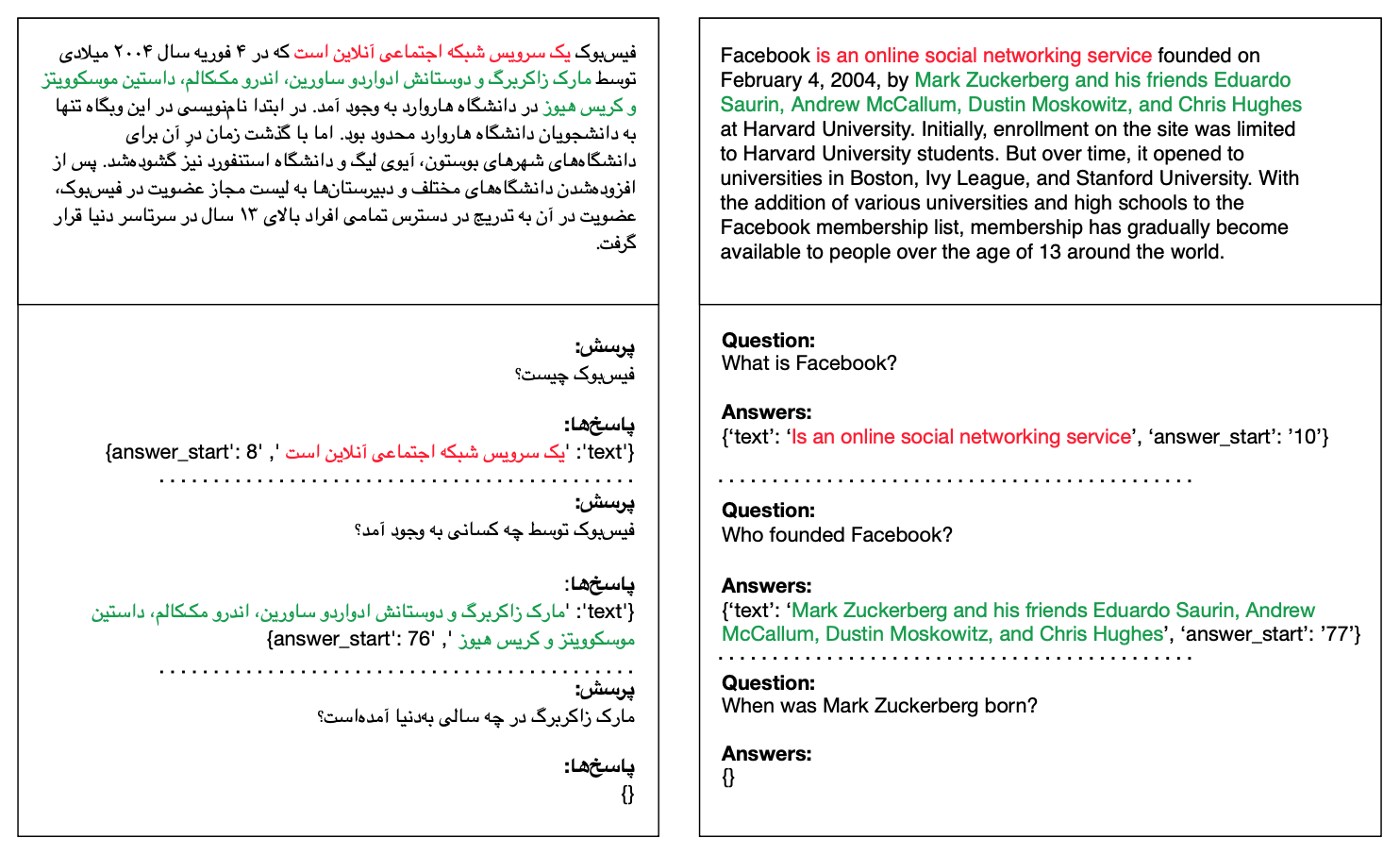}
	\caption{An example of a passage with corresponding questions. The left part is the original Persian data and the right part is the English translation of that. The third question is an example of a question with no answer. Answers are represented by the text of answer and the start token of the answer.}
	\label{fig:data-example}     
\end{figure*}

PQuAD is stored in the JSON format and consists of passages where each passage is linked to a set of questions. The unanswerable questions are marked as unanswerable and answer of the answerable questions is specified with answer's span. An example of a passage with corresponding question and answers is shown in Figure \ref{fig:data-example}.

PQuAD includes 11,000 passages and 80,000 questions where each passage includes about 7 questions. PQuAD is divided into three categories including train, validation, and test sets. A detailed statistics of PQuAD is presented in Table \ref{tab:quad-stat}.

\begin{table}[h!]
	\begin{center}
		\caption{Statistics of PQuAD dataset.}
		\label{tab:quad-stat}
		
		\begin{tabular}{  | c | c | c | c | c | }
			\hline
			&Train	& Validation 	&Test 	&Total   \\ \hline			
			Total Questions  				&63994 	&7976			&8002	&79972		 \\ \hline
			Unanswerable Questions			&15721  &1981			&1914	&19616	 \\ \hline
			Mean \# of paragraph tokens		&125  	&121			&124	&125	 \\ \hline
			Mean \# of question tokens		&10  	&11				&11		&10	 \\ \hline
			Mean \# of answer tokens		&5		&6				&5		&5		 \\ \hline
		\end{tabular}
	\end{center}
\end{table}

As mentioned before, PQuAD features a diverse range of topics. Table \ref{tab:domains} categorizes topics based on the domains associated with the articles present in the whole dataset.

\begin{table}[h!]
	\begin{center}
		\caption{Diversity of article topics present in PQuAD and sample titles of source Wikipedia pages.}
		\label{tab:domains}
		
		\begin{tabular}{  | c | c | c | }
			\hline
			Domain			& Example	& Percentage(\%)    \\ \hline				
			Person				&Pablo Picasso	&22.3	\\ \hline
			Geographical Locations		&Caspian Sea	&20.1	\\ \hline
			Science \& Tech		&Database	&6.5	\\ \hline
			Organization	&United Nations	&6.0	\\ \hline
			Sports	&Olympic Games	&5.8	\\ \hline
			Fields of Specialty	&Psychology	&5.0	\\ \hline
			Plants \& Animals	&Starfish	&4.0	\\ \hline
			Art	&Pop music	&3.3	\\ \hline
			Historical Eras	&Precambrian	&3.2	\\ \hline
			Books \& Movies	&Star Wars	&2.9	\\ \hline
			Religious	&God	&2.7	\\ \hline
			Events	&Nowruz	&2.6	\\ \hline
			Groups	&Vikings	&2.5	\\ \hline	
			Languages	&Middle Persian	&2.4	\\ \hline
			Chemistry \& Biology	&Hydrogen	&2.4	\\ \hline
			Astronomy	&Solar System	&2.2	\\ \hline	
			\,Diseases \& Medicines\,	&\,COVID-19 pandemic\,	&2.0	\\ \hline
			Objects	&Carpet	&1.6	\\ \hline	
			Others	&Immigration	&2.4	\\ \hline

			\hline					
		\end{tabular}
	\end{center}
\end{table}

\subsection{Answer types}

We categorize answer spans of answerable questions of the validation set based on their POS and NER tags, similarly to \citep{rajpurkar:2016:squad}. The results of categorization are shown in Table \ref{tab:pos}. As can be seen, the most popular types of the answers are numerical, proper noun phrase, and common noun phrase accounting for 24\%, 37\%, and 27\% of all answers, respectively. The remaining 8.5\% is distributed across adjective phrases, verb phrases, and other types.

\begin{table}[]
	\centering
	\caption{Categorization of validation set's answers based on POS and NER tags.}
	\label{tab:pos}
	\begin{tabular}{|c|c|c|}
		\hline
		POS                                 & NER           & Percentage(\%) \\ \hline
		\multirow{2}{*}{Numeric}            & Date          & 10.2  \\ \cline{2-3} 
		& Other Numeric & 13.4  \\ \hline
		\multirow{7}{*}{Proper Noun Phrase} & Person        & 14.9  \\ \cline{2-3} 
		& Location      & 12.4  \\ \cline{2-3} 
		& Group         & 2.8   \\ \cline{2-3} 
		& Organization  & 2.3   \\ \cline{2-3} 
		& Field         & 1.1   \\ \cline{2-3} 
		& Language      & 1.3   \\ \cline{2-3} 
		& Other Entity  & 3.7   \\ \hline
		Common Noun Phrase                  & -             & 27.0  \\ \hline
		Adjective Phrase                    & -             & 1.9   \\ \hline
		Verb Phrase                         & -             & 4.9   \\ \hline
		Other                               & -             & 4.1   \\ \hline
	\end{tabular}
\end{table}

\section{Experiments}
\label{sec:exp}

\subsection{Human performance}

To estimate the human performance on our PQuAD dataset, we asked a new group of crowdworkers to answer 1000 questions in the test set. These questions are from 15 randomly selected articles and all questions of each article are tasked to one of the workers. The percentage of the unanswerable questions in the samples is the same as in the dataset. Workers were informed about the existence of unanswerable questions and were asked to either mark the questions for which they found no answer or specify the answer span in the paragraph.
Performance measures used in this experiment are F1 and Exact Match (EM) metrics, the same as measures used by \citep{rajpurkar:2016:squad}. The estimated human performance on the test set is 88.3\% for F1 and 80.3\% for EM.

\subsection{Models}

We have evaluated PQuAD using two pre-trained transformer-based language models, namely ParsBERT \citep{farahani:2021parsbert} and XLM-RoBERTa \citep{conneau:2020:xlmroberta}, as well as BiDAF \citep{levy:2017bidaf} which is an attention-based model proposed for MRC.

\begin{itemize}
	\item ParsBERT: ParsBERT is a transform-based language model which uses the architecture of BERT$_{base}$ model. ParsBERT is specially developed for the Persian language and is trained on Persian texts gathered from diverse sources. \\
	
	\item XLM-RoBERTa: XLM-RoBERTa is a transformer-based language model. Similar to the mono-lingual RoBERTa language model, XLM-RoBERTa does not use the Next Sentence Prediction (NSP) task for training and it is only trained using the multilingual Masked Language Model (MLM). 
	Size of the training data is very large and it is capable of training by 100 different languages.  XLM-RoBERTa is a strong multi-lingual pre-trained language model, which has outperformed the multilingual BERT (mBEERT) on several cross-lingual tasks. \\
	
	\item BiDAF: BiDAF is a neural architecture which utilizes character-level, word-level, and contextualized embedding for text representation. Attention mechanism is used in the model for creating a query-aware representation of the input context. The model detects the span of answer from context by calculating the probability of being the begin and end token of the span for each word within the context. 
\end{itemize}

\begin{table}[!h]
	\begin{center}
		\caption{Scores of baseline models and human performance. HasAns and NoAns columns show the scores over positive and negative questions respectively.}
		\label{tab:result}
		
		\begin{tabular}{  | c | c | c | c | c | c | c | }
			\hline
			Model			& EM	& F1  	&HasAns\_EM  &HasAns\_F1  &NoAns\_EM/F1  \\ \hline			
			BNA				&54.4	&71.4	&43.9		&66.4		&87.6 	\\ \hline
			ParsBERT		&68.1	&82.0	&61.5		&79.8		&89.0 	\\ \hline
			XLM-RoBERTa		&74.8	&87.6	&69.1		&86.0		&92.7 	\\ \hline
			Human			&80.3	&88.3	&74.9		&85.6		&96.8 	\\ \hline
		\end{tabular}
	\end{center}
\end{table}

\subsection{Results}
Table \ref{tab:result} shows the performance of baseline models on the test set. According to our experiments, mono-lingual BERT underperforms multi-lingual XLM-RoBERTa which might be the result of a less effective pre-training process. Despite the close F1 scores of XLM-RoBERTa and humans, there is a 5.5\% gap between their EM scores. Based on the detailed scores of answerable and unanswerable questions, this lower performance is mostly due to the model's inability to select the shortest span containing the answer for positive questions. Using the same model, and to have a better understanding of the model's weak points, we perform an analysis on the error rates based on answer types. As reported in table\ref{tab:em_answerTyper}, the model has comparatively high performance on noun phrases and numbers. However, differences between model and human performance are more obvious in adjective and verb phrases. These types of answer spans are more variable in their structures, especially in the Persian language, and specifying the exact start and end position of them is a challenging task for the model, suggesting a direction for future works.

\begin{table}[]
	\centering
	\caption{XLM-RoBERTa and human performance on different answer categories.}
	\label{tab:em_answerTyper}
	\begin{tabular}{|c|cc|cc|}
		\hline
		\multirow{2}{*}{Answer Type} & \multicolumn{2}{c|}{XLM-RoBERTa} & \multicolumn{2}{c|}{Human}   \\ \cline{2-5} 
		& \multicolumn{1}{c|}{EM}    & F1  & \multicolumn{1}{c|}{EM} & F1 \\ \hline	
		Numeric            & \multicolumn{1}{c|}{83.51}   &92.64     & \multicolumn{1}{c|}{86.9}   &92.7    \\ \hline
		Proper Noun Phrase & \multicolumn{1}{c|}{78.4}      &88.7     & \multicolumn{1}{c|}{78.6}   &86.2    \\ \hline
		Common Noun Phrase & \multicolumn{1}{c|}{70.5}      &88.0     & \multicolumn{1}{c|}{71.7}   &84.9    \\ \hline
		Adjective Phrase   & \multicolumn{1}{c|}{57.1}      &79.5     & \multicolumn{1}{c|}{85.7}   &85.7    \\ \hline
		Verb Phrase        & \multicolumn{1}{c|}{53.3}      &83.4     & \multicolumn{1}{c|}{62.0}   &84.7    \\ \hline
		Other              & \multicolumn{1}{c|}{62.3}      &76.5     & \multicolumn{1}{c|}{68.7}   &76.4    \\ \hline
	\end{tabular}
\end{table}

\section{Conclusion}
\label{sec:conclusion}

In this paper, we present PQuAD, a Persian reading comprehension dataset consisting of 80,000 questions. This dataset is based on Wikipedia articles and its question-answer pairs are annotated by humans. Since unanswerable questions are also included in PQuAD, models are required to abstain from answering whenever no proper answer span is present in the given context. By releasing this dataset, we aim to ease research on Persian reading comprehension and the development of Persian question answering systems.



%
%

\bibliographystyle{spbasic}      
\bibliography{references}   

%
%


\end{document}